\documentclass[10pt,twocolumn,letterpaper]{article}

\usepackage[pagenumbers]{cvpr} 

\usepackage{caption,subcaption}
\usepackage{graphicx}
\usepackage{amsmath}
\usepackage{amssymb}
\usepackage{booktabs}
\usepackage{pifont}
\usepackage{fontawesome5}
\usepackage{xcolor}
\usepackage[pagebackref,breaklinks,colorlinks]{hyperref}

\usepackage[capitalize]{cleveref}
\crefname{section}{Sec.}{Secs.}
\Crefname{section}{Section}{Sections}
\Crefname{table}{Table}{Tables}
\crefname{table}{Tab.}{Tabs.}

\def\cvprPaperID{28374} 
\def\confName{CVPR}
\def\confYear{2025}
\definecolor{wilddepthblue}{HTML}{1F77B4} 

\begin{document}

\title{WildDepth: A Multimodal Dataset for 3D Wildlife Perception and Depth Estimation}
\author{%
\parbox{\textwidth}{\centering
Muhammad Aamir\textsuperscript{1}\thanks{Equal contribution}\quad
Naoya Muramatsu\textsuperscript{2}\footnotemark[1]\quad
Sangyun Shin\textsuperscript{1}\footnotemark[1]\quad
Matthew Wijers\textsuperscript{1}\\[0.6em]
Jia-Xing Zhong\textsuperscript{1}\quad
Xinyu Hou\textsuperscript{1}\quad
Amir Patel\textsuperscript{3}\quad
Andrew Loveridge\textsuperscript{1}\quad
Andrew Markham\textsuperscript{1}\\[0.9em]
\textsuperscript{1}University of Oxford\quad
\textsuperscript{2}University of Cape Town\quad
\textsuperscript{3}University College London\\[0.6em]
{\small\protect\texttt{\{muhammad.aamir,sangyun.shin,jiaxing.zhong,xinyu.hou,andrew.markham\}@cs.ox.ac.uk}\\
\protect\texttt{matthew.wijers@biology.ox.ac.uk, andrew.loveridge@lmh.ox.ac.uk, mrmnao001@myuct.ac.za, amir.patel@ucl.ac.uk}}\\[0.5em]
\protect\href{https://yunshin.github.io/WildDepth/}\protect\href{https://yunshin.github.io/WildDepth/}{%
  \protect\textcolor{wilddepthblue}{\protect\underline{\protect\textbf{\faGlobe\ Project Page}}}%
}
}%
}

\maketitle
\begin{abstract}
Depth estimation and 3D reconstruction have been extensively studied as core topics in computer vision. Starting from rigid objects with relatively simple geometric shapes, such as vehicles, the research has expanded to address general objects, including challenging deformable objects, such as humans and animals. However, for the animal, in particular, the majority of existing models are trained based on datasets without metric scale, which can help validate image-only models. To address this limitation, we present \textbf{\textit{WildDepth}}, a multimodal dataset and benchmark suite for depth estimation, behavior detection, and 3D reconstruction from diverse categories of animals ranging from domestic to wild environments with synchronized RGB and LiDAR. Experimental results show that the use of multi-modal data improves depth reliability by up to 10\% RMSE, while RGB–LiDAR fusion enhances 3D reconstruction fidelity by 12\% in Chamfer distance. By releasing WildDepth and its benchmarks, we aim to foster robust multimodal perception systems that generalize across domains.


\end{abstract}

\section{Introduction}
\label{sec:intro}
\begin{figure}[htbp]
    \centering
    \includegraphics[width=\linewidth]{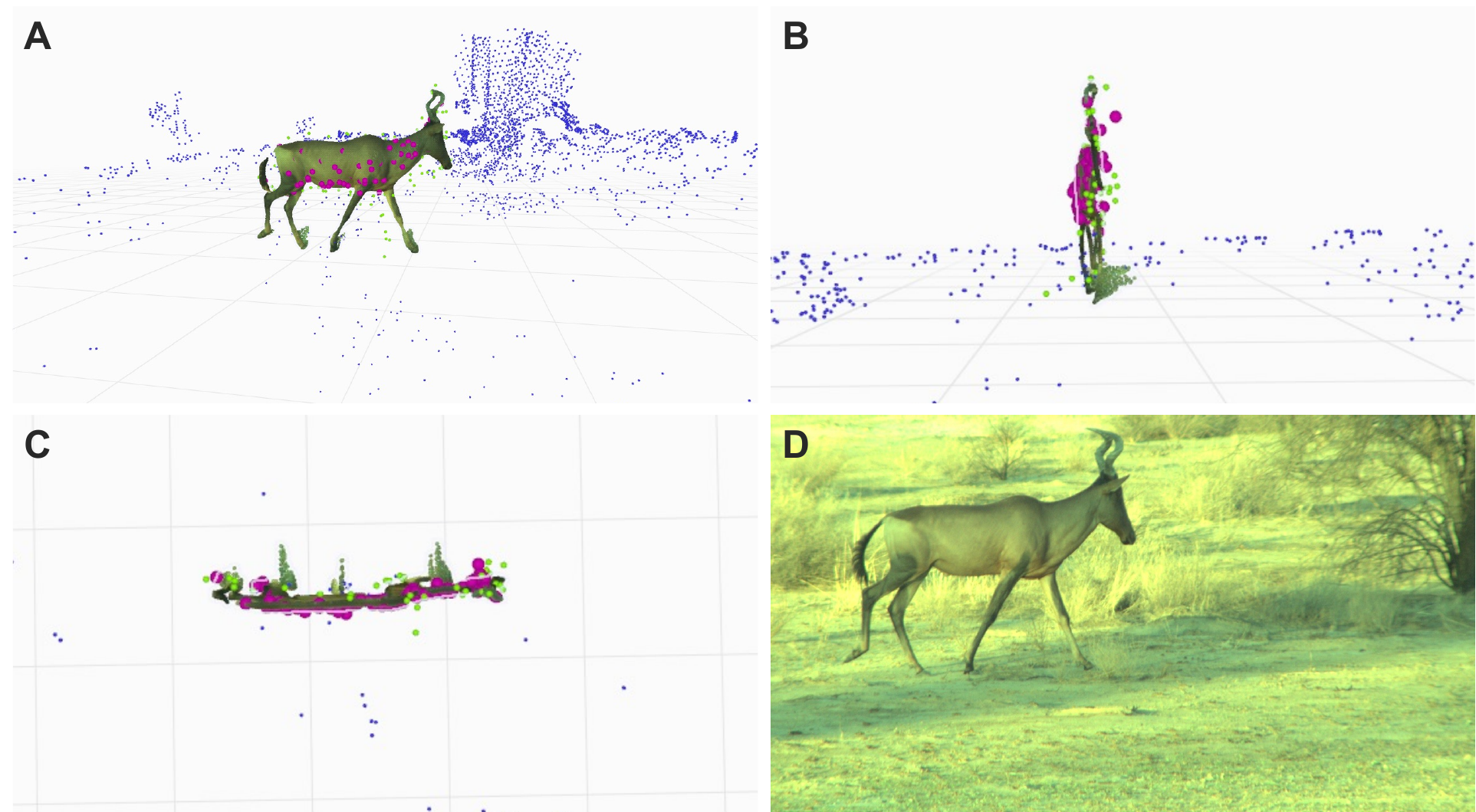}
    \caption{Reconstruction result for a walking Red hartebeest.}
    \label{fig:qualitative_results}
    \vspace{-5mm}
\end{figure}

Monitoring wildlife behavior and population dynamics has become increasingly crucial as biodiversity faces unprecedented threats from habitat loss, poaching, and climate change. Recent advances in computer vision and remote sensing have opened new possibilities for non-invasive monitoring at scale, enabling the observation of species in their natural environments without human interference. From camera traps to aerial surveys, multimodal data have become fundamental in understanding species distribution, morphology, and ecological interactions. However, the development of robust models for wildlife monitoring remains constrained by the scarcity of comprehensive, high-quality, and multimodal field datasets. Most existing datasets are limited in modality, temporal scope, or geographical diversity, leaving a significant gap in the training and benchmarking of deep learning methods for ecological applications \cite{binley2024making}.

\begin{table*}[htbp]
\centering
\caption{Publicly available datasets for wildlife monitoring}
    \begin{tabular}{lcccc}
    \toprule
    \textbf{Datasets} & \textbf{Source} & \textbf{\# Species} & \textbf{\# Videos/Images} & \textbf{MultiModal}  \\
    \midrule
    AnimalKingdom \cite{ng2022animal} & YouTube & 850 & 30k & \ding{55}\\
    MammalNet \cite{chen2023mammalnet}&  YouTube & 173 & 20k & \ding{55}\\
    KABR \cite{kholiavchenko2024kabr}&  Drone & 3& 13k & \ding{55}\\
    Meerkats \cite{rogers2023meerkat} & Zoo & 1 & 35 & \ding{55}\\
    ChimpACT \cite{ma2023chimpact}& Zoo &1& 163 &\ding{55}\\
    MammAlps \cite{gabeff2025mammalps}& Camera Traps & 5& 6k & \ding{55}\\
    LoTE \cite{liu2023lote}& Camera Traps & 11& 10k &\ding{55} \\
    BaboonLand \cite{duporge2025baboonland} & Drone & 1 & 30k & \ding{55}\\
    \textbf{WildDepth}& Safari Park + Wild & 29 & 202k &\ding{51} \\
    \bottomrule
    \end{tabular}
\label{tab:literature}
\end{table*}
To address this gap, we present \textit{WildDepth}, a unified multimodal wildlife dataset comprising three subsets collected in real-world environments that bridge the disciplines of ecology, computer vision, and artificial intelligence.
The first subset was collected in the Kgalagadi Transfrontier Park, South Africa. A combination of RGB-LiDAR recordings using a zoom-lens camera setup, collected a wide range of wildlife data from apex predators such as lions, cheetahs and leopards to large herbivores such as springbok and gemsbok.
The second subset was recorded across 4 days in June 2025 in the Buby Valley Conservancy, Zimbabwe, combining LiDAR and RGB  data under various natural illumination and environmental conditions.
The third subset was collected in July 2025 at the Longleat Safari Park, United Kingdom, using a suite of multimodal sensors that included RGB cameras, GoPro Hero 11, and LiDAR systems that captured synchronized data.
Together, these three subsets form WildDepth, one of the first multimodal field collections designed explicitly for wildlife image processing tasks such as pose estimation, 3D reconstruction, morphological analysis and depth prediction.

The objective of assembling these datasets is two-fold.
First, they provide rich multimodal ground truth for developing AI systems capable of fine-grained spatial and behavioral understanding of wildlife.
Second, they serve as benchmark datasets to facilitate reproducible research across various conservation and ecological monitoring tasks.
Each modality contributes complementary information, for example, LiDAR aids in accurate depth and morphological measurements \cite{andualem2024channel}.
The synchronized collection of RGB-LiDAR supports cross-modal learning, enabling models to infer structure and behavior from incomplete or noisy data.

In the broader context of wildlife conservation, integration of such datasets into AI workflows aligns with a growing recognition that traditional observation-based methods are insufficient for large scale ecosystem monitoring.
Initiatives like GBIF~\cite{lane2007global}, eBird~\cite{sullivan2009ebird}, and the IUCN Red List~\cite{list2011iucn} have demonstrated the importance of open, standardized data sharing, but most rely on 2D or textual information, which limits their applicability to 3D and behavioral modeling.
By providing high resolution, multimodal sensor data, the proposed datasets offer new opportunities for advancing computational ecology, data-driven conservation, and automated biodiversity assessment.
This work draws inspiration from recent efforts in multimodal wildlife behavior datasets, such as MammAlps~\cite{gabeff2025mammalps}, which demonstrated the potential of integrating video, audio, and segmentation data for multi-view animal behavior analysis.
However, our datasets differ in scope and modality, emphasizing 3D perception and cross-domain sensor fusion rather than camera-trap based video analytics.
Through experiments on diverse wildlife monitoring tasks including depth and morphology estimation, pose tracking and 3D reconstruction, we demonstrate how multimodal sensor fusion enhances model robustness and ecological interpretability.

Our contributions can be summarized as follows:
\begin{itemize}
    \item WildDepth, a unified multimodal wildlife dataset with three subsets collected in ecologically distinct environments using synronised RGB and LiDAR imaging systems.
    \item A set of benchmark experiments across depth estimation, pose detection, 3D reconstruction, and population density estimation tasks, establishing baselines for AI-driven conservation research.
    \item Insights into multimodal learning for ecological data, showcasing how combining heterogeneous sensors can improve accuracy and generalization in field-based wildlife monitoring.
\end{itemize}
\section{Related Work}
Recent years have witnessed a convergence between computer vision and ecology, driven by the demand for vision models capable of understanding general objects across domains in wild.


\noindent\textbf{Vision-based Wildlife Monitoring.}
Traditional ecological monitoring relies heavily on camera traps, aerial imagery, or manual observation. These methods have inspired a series of computer vision applications for animal detection, identification, and behavioral tagging~\cite{ravoor2020deep}.
Datasets such as MammAlps~\cite{gabeff2025mammalps}, LoTE~\cite{liu2023lote}, and BaboonLand~\cite{duporge2025baboonland} have provided essential benchmarks for video-based behavior recognition, yet most are limited to RGB or audio inputs and lack geometric or metric-scale calibration. As a result, these datasets mainly support classification rather than quantitative 3D reasoning.
Furthermore, the majority of existing animals pose and meshing models are derived from internet-based video and image datasets for example, Animal Kingdom~\cite{ng2022animal} and MammalNet~\cite{chen2023mammalnet}, which capture species appearances but lack metric ground truth, which cannot be quantitatively validated for 3D accuracy required for physical realism. 

\noindent\textbf{Monocular Depth Estimation and 3D Understanding.}
Monocular Depth Estimation (MDE) has become a central topic in computer vision, progressing from classical stereo geometry toward self-supervised and transformer based learning.
Models such as MonoDepth~\cite{li2024improved}, DPT~\cite{ranftl2021vision}, DepthPro~\cite{bochkovskii2024depth}, and DepthAnything~\cite{yang2024depth} have achieved remarkable results in structured indoor or urban environments (e.g., KITTI~\cite{liao2022kitti}, NYUv2~\cite{silberman2012indoor}).
However, these benchmarks provide stable geometry, controlled lighting, and static objects where the conditions rarely met in wild habitats.
In contrast, wildlife scenes are highly unstructured, dominated by irregular terrain, vegetation, and non-rigid subjects that move across large depth ranges.
Metric scale ground truth from such environments is insufficient, which precludes rigorous evaluation of MDEs under realistic ecological conditions.
The WildDepth dataset introduces precisely aligned RGB-LiDAR pairs in natural outdoor environments, offering one of the few sources of metric depth for small and distant targets, where monocular methods typically fail.

\noindent\textbf{3D Reconstruction and 4D Deformable Modeling.}
The use computer vision algorithms in 3D reconstruction of objects including animals in motion extends beyond static depth prediction into the domain of deformable 4D perception, where geometry, texture, and time must be jointly modeled.
While human motion capture and synthetic datasets (e.g., AMASS~\cite{mahmood2019amass}, Human3.6M~\cite{ionescu2013human3}) have enabled rapid progress in human body modeling, equivalent resources for wild animals remain limited.
Existing research on animal meshing or skeleton recovery (e.g., SuperAnimal~\cite{ye2024superanimal}, ChimpACT~\cite{ma2023chimpact}) are trained on video data without metric calibration, preventing accurate validation of predicted poses or surface deformations.
In such circumstances, the need of a dataset that provides synchronized 3D data of freely moving, deformable subjects in natural habitats is important.

\noindent\textbf{Multimodal and Domain Robust Perception.}
The computer vision community has increasingly recognized the value of multimodal learning, combining RGB, LiDAR and depth information to enhance perception in challenging environments~\cite{xiao2020audiovisual}.
Yet, most multimodal datasets are still urban or indoor in nature, focusing on autonomous driving or robotics (e.g., KITTI-360~\cite{liao2022kitti}, nuScenes~\cite{caesar2020nuscenes}).
The ecological domain remains an outlier, it combines large dynamic ranges, deformable subjects, and environmental interference that can degrade each modality independently.

By integrating RGB–LiDAR sensing across three geographically and ecologically distinct sites, the \textit{WildDepth} dataset enables the study of cross-domain robustness and perception models transfer between grasslands, and semi-controlled environments.
This aspect is especially valuable for designing foundation models capable of adapting to unseen environmental conditions.

\section{WildDepth}

\subsection{Data Collection}

WildDepth comprises three subsets collected in 2022 and 2025 from semi-controlled safari park settings and natural, wild habitats, providing a diverse representation of animal appearances and environmental conditions.

\paragraph{Kgalagadi Transfrontier Park, South Africa}
The first dataset component was captured in December 2022 at the Kgalagadi Transfrontier Park, South Africa.
This data was collected using the WildPose system~\cite{muramatsuWildPoseLongRange3D2024}, a long-range 3D motion capture system combining a zoom-lens camera and LiDAR.
The system was mounted on a vehicle door for a 13-day survey along fixed routes near the Two Rivers and Nossob camp sites.
We recorded synchronized RGB-LiDAR sequences of land animals and large birds, with capture durations ranging from 30 to 900 seconds.
This data features a wide range of wildlife, including apex predators like lions and cheetahs, as well as large herbivores such as springbok and gemsbok.
To ensure data quality under varying field conditions, we maintained a high shutter speed ($1/1000\,\mathrm{s}$) and actively managed the lens aperture (e.g., f/11--32) to balance adequate image brightness with sufficient depth of field.

\paragraph{Zimbabwe Wildlife Conservancy}
The second wild dataset was collected in June 2025 in the Bubye Valley Conservancy, a private wildlife area in southern Zimbabwe dominated by mopane and acacia woodland habitats.
We built a custom data collection rig comprising a 4K RGB camera linked with LiDAR (Livox Mid-70, DJI).
Data were recorded using both a mobile setup, with the rig mounted on a vehicle window, and a stationary setup, where the rig was positioned near animals as they moved and grazed.
We collected over 2 hours (~84,700 frames) of synchronized RGB-LiDAR recordings of 6 species (giraffe, zebra, warthog, donkey, goat, and plover birds).
This dataset presents natural challenges including changing illumination, occlusions, and long-range motion, making it useful for assessing model robustness in real-world settings.

\paragraph{Longleat Safari Park, United Kingdom}
To complement the wild datasets, we collected an additional dataset at Longleat Safari Park in Wiltshire, UK, which hosts a wide variety of habituated animal species in spacious, open enclosures with relatively sparse vegetation.
This setting allowed us to record a broader range of species under more controlled conditions and with lower scene complexity compared to wild environments. The data collection rig consisted of a RGB camera paired with a LiDAR (Livox Avia, DJI).
Similar to our data collection in Zimbabwe, data were recorded using both a mobile setup, with the rig mounted on a vehicle window, and a stationary setup, where the rig was positioned on a tripod on a viewing platform overlooking an enclosure.
In total, we recorded 1.25 hours of RGB-LiDAR data comprised of 70 videos (approximately 42500 frames) of different species (e.g. lion, tiger, cheetah, giraffe, ostrich, zebra, wolf, ox, elephant, rhino).

Across all three environments, the data were collected with precise temporal synchronization ($\pm10\,\mathrm{ms}$ tolerance) between the RGB and LiDAR streams using GPS pulse-per-second (PPS) triggers.
The combined dataset covers diverse species morphologies, ecological contexts, and terrain structures, producing dense geometric ground truth and a wide range of appearance variations.
Together, these three subsets form WildDepth, a comprehensive multimodal wildlife dataset that supports the development and evaluation of perception models for depth estimation, 3D reconstruction, and behavioral understanding.
The multi habitat data ensure that models trained on this dataset can generalize across distinct natural domains, from wild savannahs and woodlands to semi-controlled enclosures, advancing the goal of scalable and robust computer vision for wildlife conservation.


\subsection{Data Processing}
We utilize Robot Operating System (ROS)~\cite{macenski2022robot} as our framework for collecting data across different environments described above.
All the collected data is processed for use in benchmarks, such as depth estimation and 3D reconstruction, which consist of calibrated pairs of RGB images and LiDAR scans.
These provide a metric ground truth of depth for each task. Since the LiDAR scans at 10 frames per second (FPS), and the camera operates at 30 FPS, the synchronization between the camera and LiDAR is based on finding the closest matching frames of the camera for each LiDAR scan, based on their timestamps.

\subsection{Data annotation}

A semi-automated annotation workflow was developed to efficiently label large scale multimodal data.
The process involved two main phases, object detection and tracking, and behavioral labeling.
Initial animal detections were obtained using MegaDetector~\cite{beeryMegadetector2019} fine-tuned on preliminary RGB samples.
For 3D modalities (LiDAR), detections were projected into the 2D camera frame via calibrated extrinsics, and temporal consistency was implemented using ByteTrack~\cite{zhang2022bytetrack}.
These detections generated continuous tracks representing individual animals across frames and modalities.
In the second phase, each frame was manually verified and annotated for species behavior (e.g., walking, grazing, resting).
LiDAR point clouds were used to measure body shape and relative dimensions. Please refer to the supplementary material for further details on data annotation.

\section{Experiment}
\begin{figure*}[htbp]
\addtolength{\tabcolsep}{-0.4em}
\begin{tabular}{cccc}
\captionsetup{width=.8\linewidth}
\centering

     Frame & DPT & DepthPro & DepthAnything \\

    \subfloat{\includegraphics[width = 1.7in]{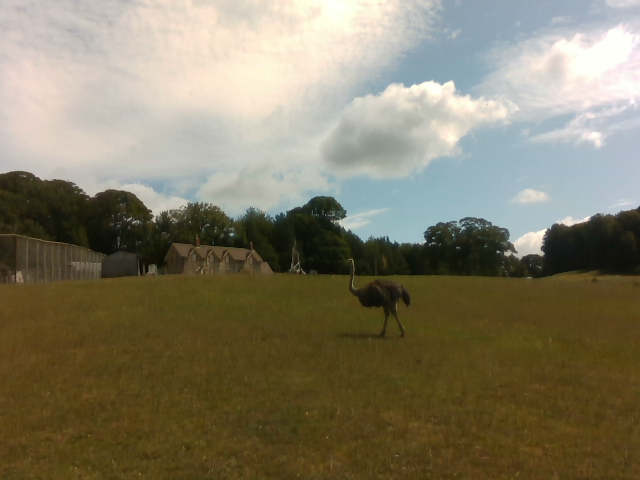}} &
        \subfloat{\includegraphics[width = 1.7in]{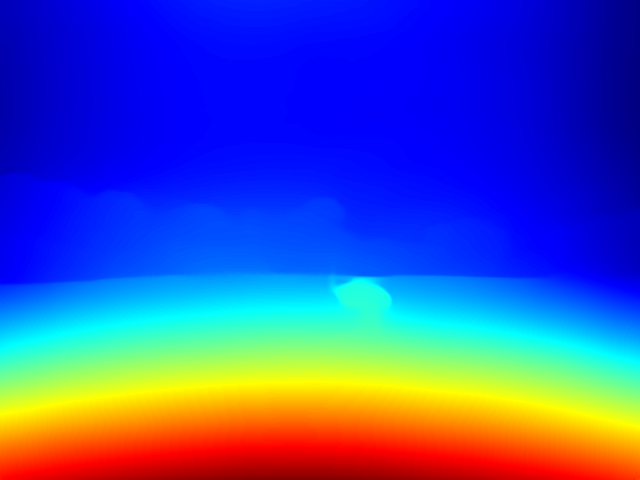}} &
        \subfloat{\includegraphics[width = 1.7in]{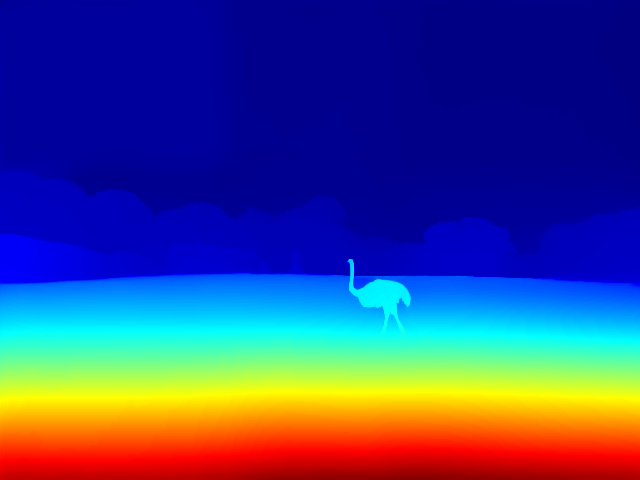}} &
        \subfloat{\includegraphics[width = 1.7in]{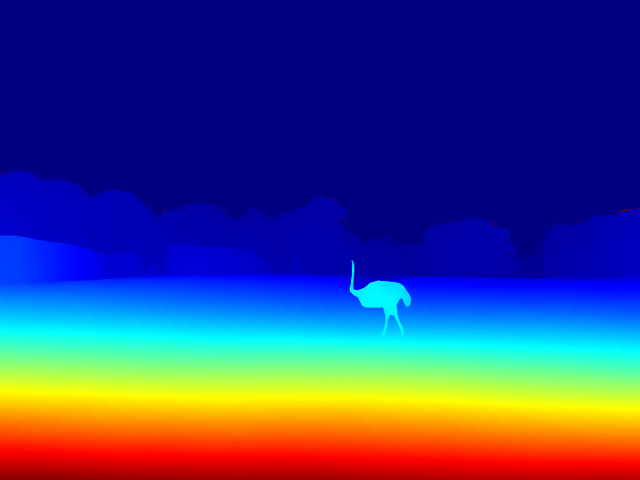}}\\

     \subfloat{\includegraphics[width = 1.7in]{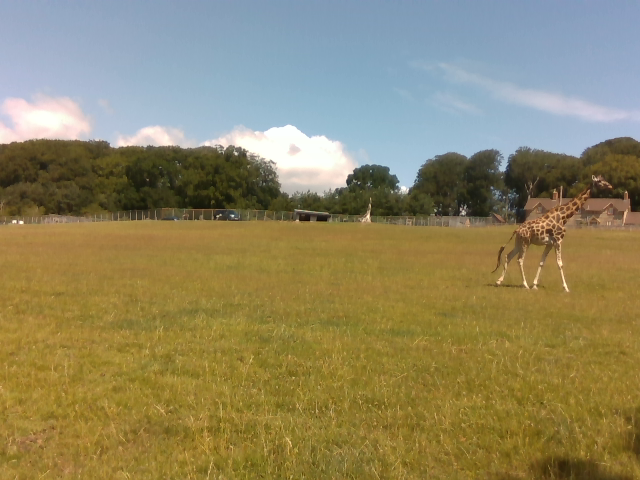}} &
        \subfloat{\includegraphics[width = 1.7in]{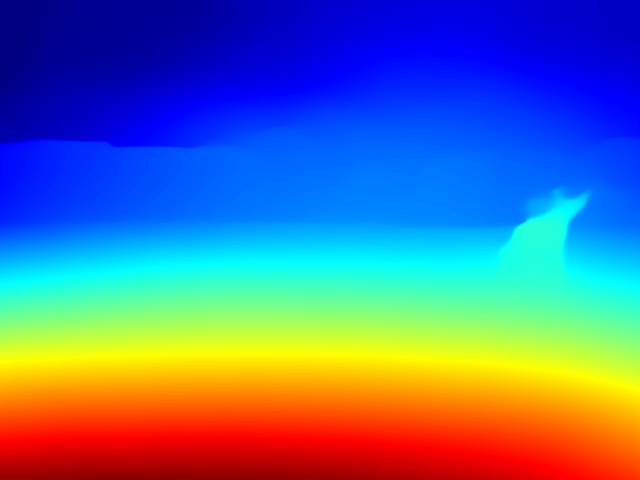}} &
        \subfloat{\includegraphics[width = 1.7in]{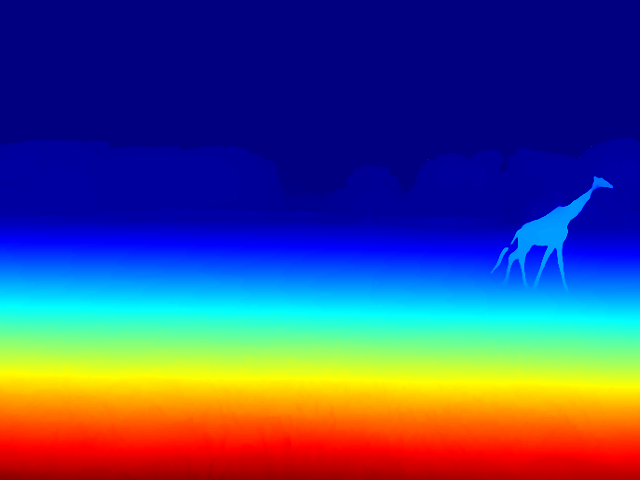}} &
        \subfloat{\includegraphics[width = 1.7in]{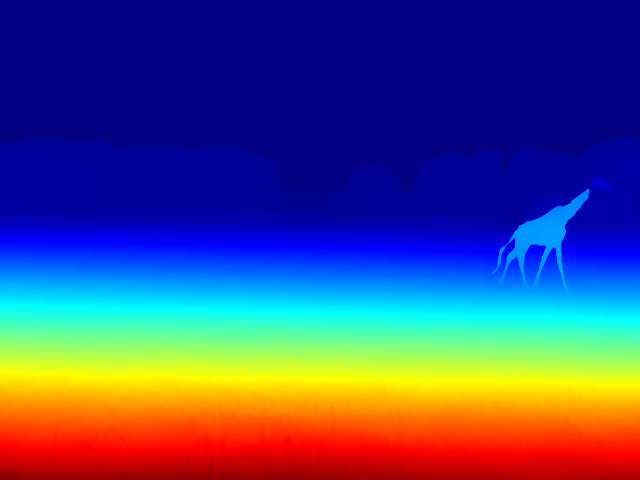}} \\

     \subfloat{\includegraphics[width = 1.7in]{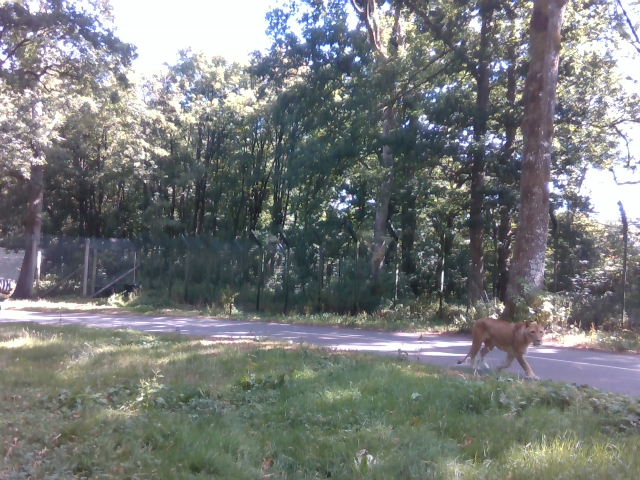}} &
        \subfloat{\includegraphics[width = 1.7in]{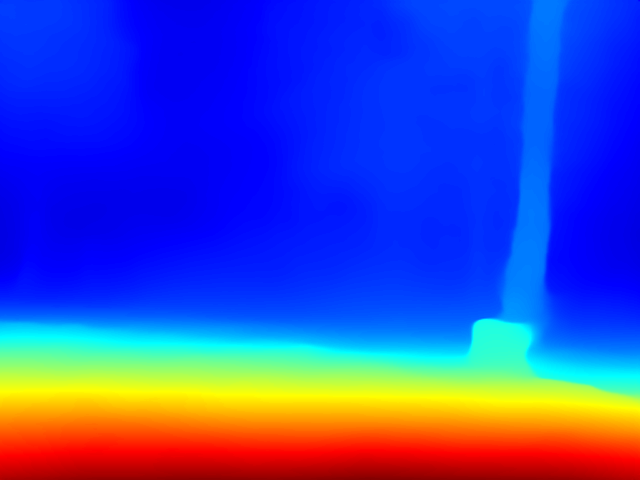}} &
        \subfloat{\includegraphics[width = 1.7in]{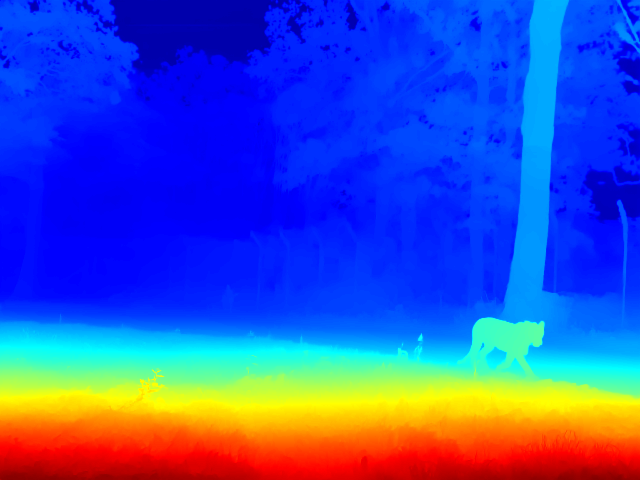}} &
        \subfloat{\includegraphics[width = 1.7in]{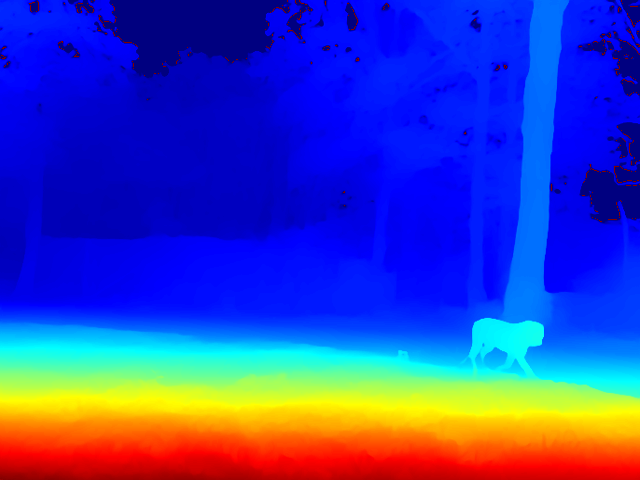}}
\end{tabular}
 
\caption{Depth Maps by different MDEs for randomly selected frames from our Dataset}
\label{fig:depthmaps}
\end{figure*}

\begin{figure}[htbp]
\centering
\includegraphics[width=\linewidth]{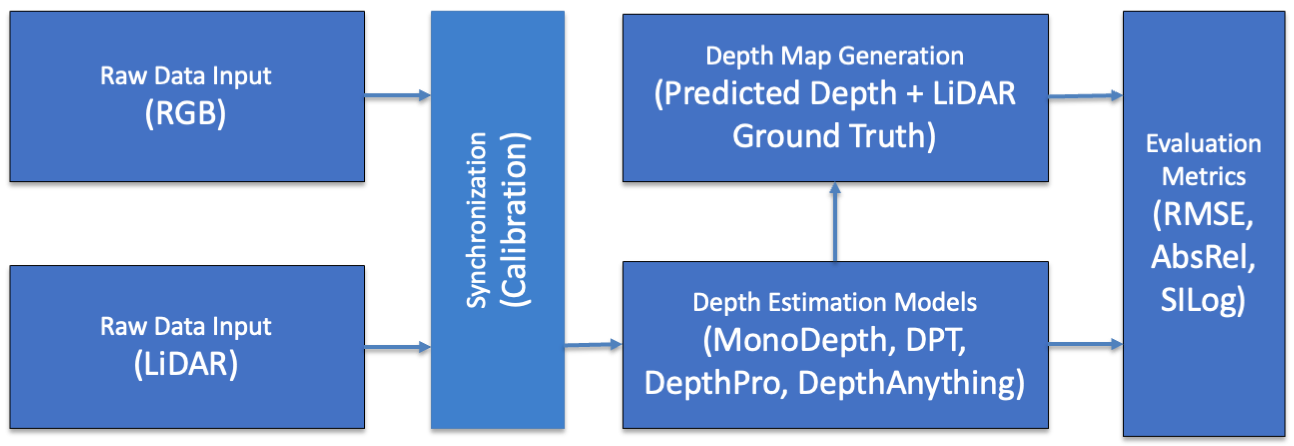}
\caption{Depth Estimation Pipeline}
\label{fig:depth_est_pipeline}
\end{figure}
This section presents the experimental evaluation conducted on the three subsets of WildDepth: Longleat Safari Park (UK), Kgalagadi Transfrontier Park (South Africa), and Bubye Valley Conservancy (Zimbabwe).
Our experiments target three key computer vision and ecological tasks: distance estimation, pose and behavior detection, and 3D reconstruction.
Each task was designed to evaluate different aspects of the utility of multimodal data and to establish baselines for future research in wildlife-oriented perception and reconstruction.
All experiments were implemented in PyTorch, trained with mixed precision on NVIDIA 2080Ti GPUs.
Camera and LiDAR modalities were used for ground truth.

\subsection{Depth Estimation}

Accurate distance estimation in wildlife imagery is a critical prerequisite for depth-aware behavioral analysis, morphometrics and population abundance estimation (distance sampling).
Traditional stereo or structured-light approaches are often impractical in dynamic, open environments where sensor placement and animal motion are uncontrolled.
Therefore, monocular depth estimation models, pretrained on large scale synthetic or real-world datasets, were evaluated and adapted for wildlife scenes.
Figure~\ref{fig:depth_est_pipeline} represents the pipeline for depth estimation.

We setup and benchmarked four state-of-the-art depth estimation models:
\begin{itemize}
    \item DepthAnything \cite{yang2024depth} -- a large-scale self-supervised model using unlabeled visual data for universal depth prediction.
    \item DPT \cite{ranftl2021vision} -- the Dense Prediction Transformer, a robust encoder-decoder architecture that transfers from vision transformers (ViT) for fine-grained depth recovery.
    \item DepthPro \cite{bochkovskii2024depth} -- a diffusion-inspired depth model optimized for geometric consistency.
    \item MonoDepth2 \cite{li2024improved} -- a self-supervised photometric model that learns depth from stereo image pairs and monocular sequences without requiring ground truth.
\end{itemize}
In all experiments, we used LiDAR depth maps as ground truth. For evaluation, we computed the RMSE, absolute relative error (AbsRel), and scale-invariant log error (SILog) in the test.

The results represented in Table~\ref{tab:depth_comparison}, DepthAnything achieved the best overall performance across all evaluation metrics, reducing RMSE by ~23\% compared to MonoDepth2 and outperforming the next-best model (DepthPro) by a relative ~20\%.

Qualitative depth visualizations (Figure~\ref{fig:depthmaps}) confirm these observations: DepthAnything preserved smooth terrain gradients and correctly delineated body contours even under varying illumination, while DepthPro excelled at maintaining fine structural edges near boundaries (e.g., horns, legs or tails).
DPT captured the scene layout coherently, but occasionally blurred small-scale features.

\begin{table*}[htbp]
\centering
    \caption{Depth estimation performance comparison. Lower is better for RMSE, AbsRel, and SILog; higher is better for $\delta < 1.25$.}
    \begin{tabular}{lcccc}
    \toprule
    \textbf{Model} & \textbf{RMSE} $\downarrow$ & \textbf{AbsRel} $\downarrow$ & \textbf{SILog} $\downarrow$ & $\boldsymbol{\delta < 1.25}$ $\uparrow$ \\
    \midrule
    MonoDepth         & 2.52 & 0.213 & 0.317 & 0.76 \\
    DPT               & 2.42 & 0.192 & 0.301 & 0.80 \\
    DepthPro          & 2.18 & 0.176 & 0.278 & 0.83 \\
    DepthAnything     & 1.94 & 0.161 & 0.254 & 0.86 \\
    \bottomrule
\end{tabular}
\label{tab:depth_comparison}
\end{table*}

\subsection{Behavior Recognition}
\label{sec:behavior}
\begin{figure}[h]
        \subfloat[Lioness]{%
            \includegraphics[width=.5\linewidth,height=1.35in]{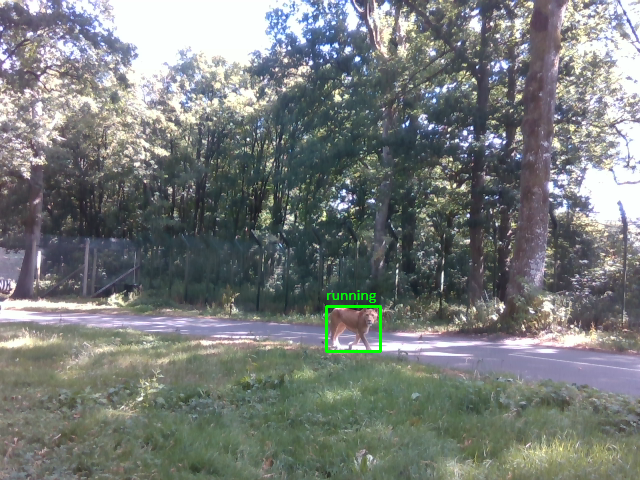}%
            \label{subfig:a}%
        }
        \subfloat[Lion]{%
            \includegraphics[width=.5\linewidth,height=1.35in]{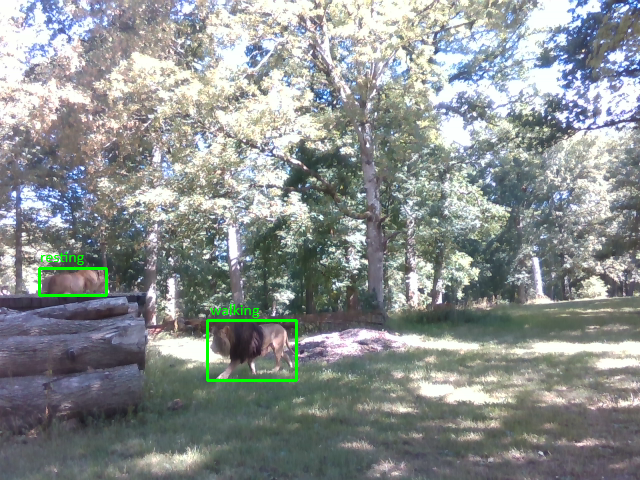}%
            \label{subfig:b}%
        }\\
        \subfloat[Donkey]{%
            \includegraphics[width=.5\linewidth,height=1.35in]{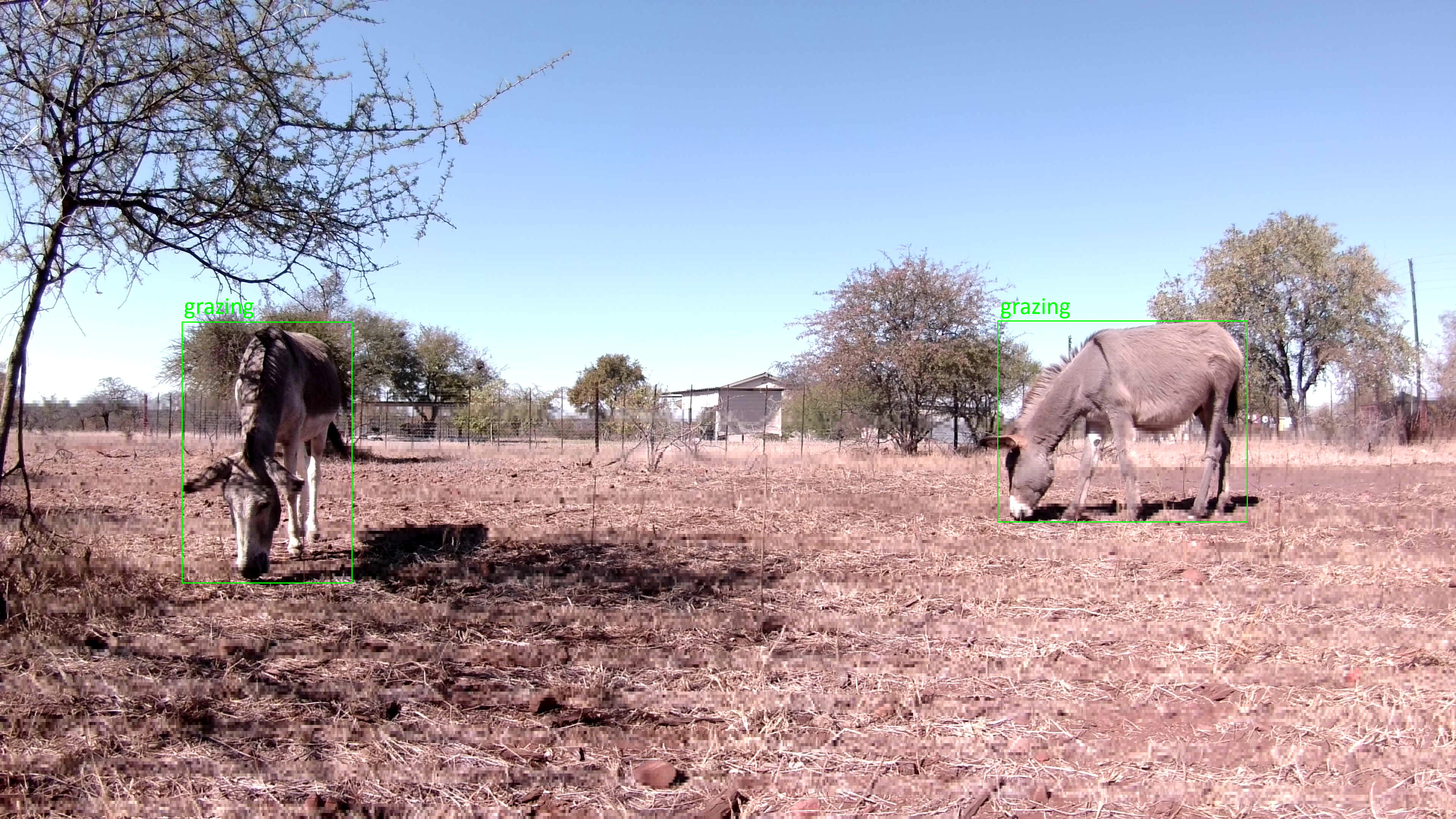}%
            \label{subfig:c}%
        }
        \subfloat[Ankole cow]{%
            \includegraphics[width=.5\linewidth,height=1.35in]{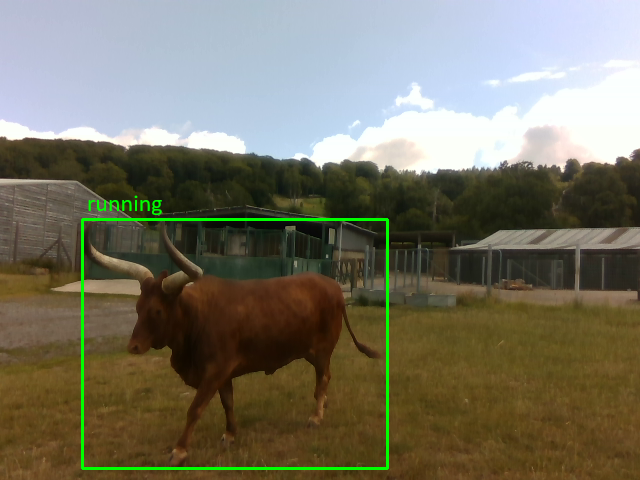}%
            \label{subfig:f}%
        }
    \caption{Behaviour recognition for randomly selected frames from our WildDepth dataset. 
    }
    \label{fig:behavior}
\end{figure}

For behavior detection, we aim to identify naturalistic actions from unconstrained field imagery. Unlike controlled environments, our sequences exhibit large appearance variations, occlusions, and motion blur due to natural lighting and unpredictable animal movements. To address these challenges, we adopt a spatiotemporal modeling approach that builds on recent advances in video-based behavior recognition. We train transformer-based classifiers following the architecture of VideoMAE~\cite{wang2023videomae} using RGB to recognize 12 activities (e.g., walking, resting, chasing) and 20 fine-grained actions (e.g., head-turning, grooming, grazing). Each video clip was 5-10 seconds long ($\approx$150 frames) and paired with the corresponding Lidar motion signatures. A few examples from our dataset are given in Figure~\ref{fig:behavior}.

\begin{figure}[htbp]
\centering
\includegraphics[width=\linewidth]{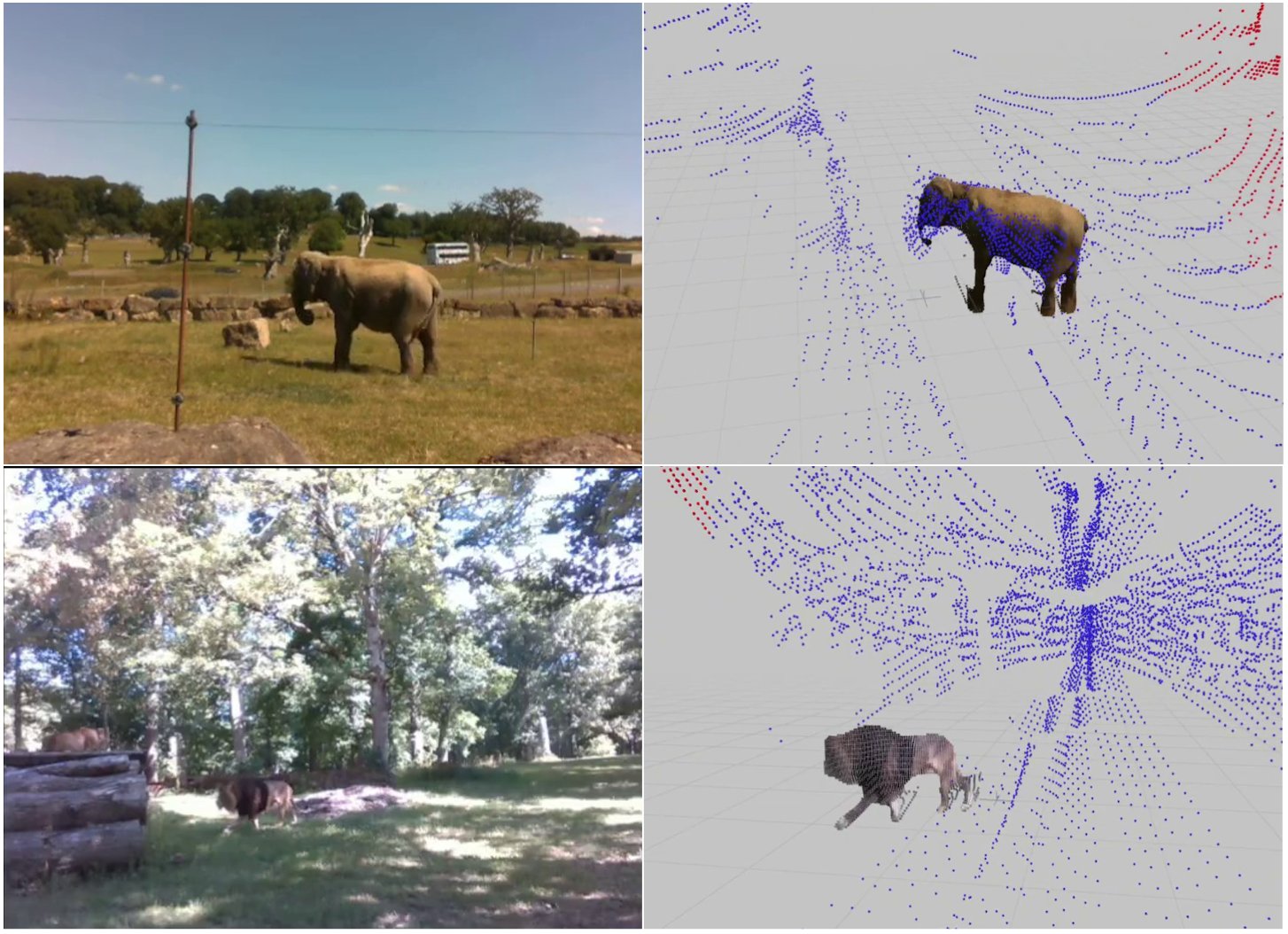}
\caption{Examples from target point densification of walking Elephant (top) and Lion (bottom).}
\label{fig:examples_tpd}
\vspace{-4mm}
\end{figure}

\subsection{3D reconstruction}
\label{sec:3dreconstruction}

\begin{figure*}[htbp]
\centering
\includegraphics[width=0.8\linewidth]{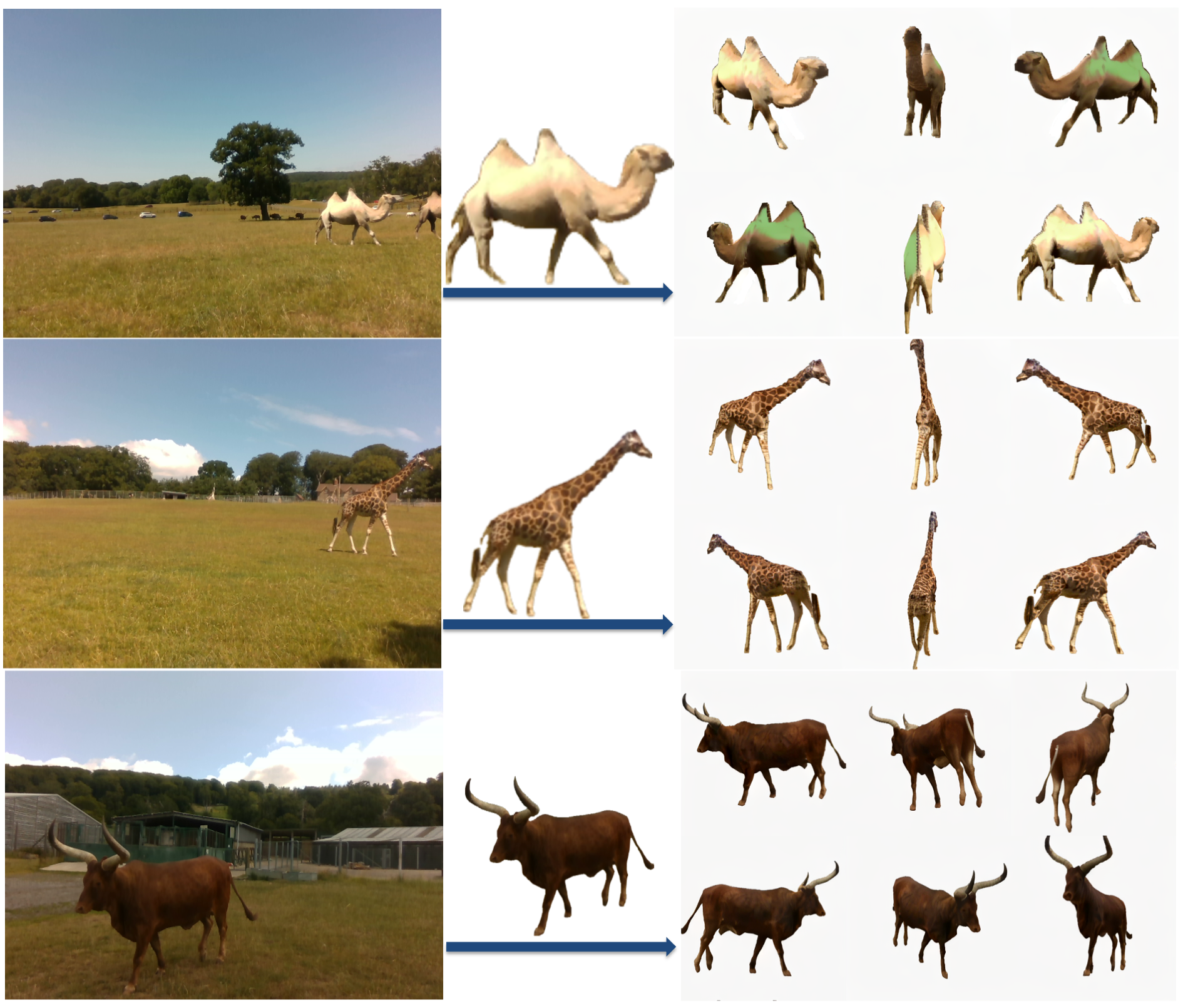}
\caption{3D reconstructions of walking Camel (top), Giraffe (middle), and Ankole cow (bottom).}
\label{fig:examples_3DR}
\end{figure*}

To reconstruct the animals 3D form, for each video, we first isolate the target animal in every frame using SAM2~\cite{raviSAM2Segment2024}, which provides high-quality segmentation masks.
These masks are used to extract and stabilize the foreground region across time, effectively reducing background clutter and motion inconsistencies. The resulting masked monocular clips serve as clean inputs for 4D reconstruction.
We then initialize a canonical set of 3D Gaussians in a reference space and employ Gaussian-Splash~\cite{feng2025gaussian}, a Gaussian-splatting–based framework, to optimize a time-dependent deformation field. This field continuously warps the canonical representation to match the observed frames, capturing both fine geometric details and temporal dynamics.
Through this process, we obtain temporally consistent 4D reconstructions that preserve the appearance, motion, and structure of the animal across the sequence.
The combination of SAM2 based segmentation and Gaussian-Splash reconstruction enables robust modeling even under complex backgrounds and nonrigid motion typical of real-world animal videos.
Figure~\ref{fig:examples_3DR} shows random frames and it reconstruction examples.

\subsection{Target Point Densification}
\label{sec:densification}

To overcome the extreme sparsity of LiDAR data for long-range wildlife, we perform a case study in which sparse, metrically accurate LiDAR returns are fused with dense, but scale-ambiguous, depth maps from a monocular depth estimator (MDE).
Rather than introducing a new algorithm, our goal is to illustrate how WildDepth can support RGB-LiDAR fusion methods for metric depth correction in realistic field conditions.
We treat this densification step as one depth-correction baseline built on top of standard monocular depth estimation.

We follow a pipeline based on prior work on long-range wildlife 3D reconstruction~\cite{muramatsuWildPoseLongRange3D2024} and subsequent thesis extensions.
A state-of-the-art MDE (MoGe2~\cite{wangMoGe2AccurateMonocular2025}) predicts the scene's relative geometry from a single RGB frame, while a segmentation model (SAM2~\cite{raviSAM2Segment2024}) generates a precise animal mask.
This mask is used to filter the raw LiDAR point cloud, isolating a sparse set of metrically accurate ``animal points'' on the target.

To increase robustness, we include two filtering steps before alignment.
First, to mitigate MDE artefacts common in telephoto imagery (e.g., ``cliff-edge'' transitions from slight defocus), we apply a Sobel-based boundary filter to the MDE output, removing unreliable depth estimates near strong gradients.
Second, we employ a RANSAC-based Scale-Shift Alignment (SSA) algorithm that uses the filtered animal points as metric anchors to solve for a global affine transformation (scale and shift) that projects the dense MDE geometry into metric space.

We evaluate this baseline on a challenging real-world sequence of a walking red hartebeest at approximately 97\,m, captured with the WildPose system~\cite{muramatsuWildPoseLongRange3D2024}.
To quantify performance, we measure the Mean Absolute Error (MAE) and the coefficient of determination ($R^2$) between the final metric depth map and the ground-truth LiDAR measurements.
As a simple but informative baseline, we also consider a Planar Depth model that approximates the animal's surface with a single 3D plane, fitted via RANSAC to the sparse LiDAR points within the animal's mask.

\begin{figure}[htbp]
    \centering
    \includegraphics[width=\linewidth]{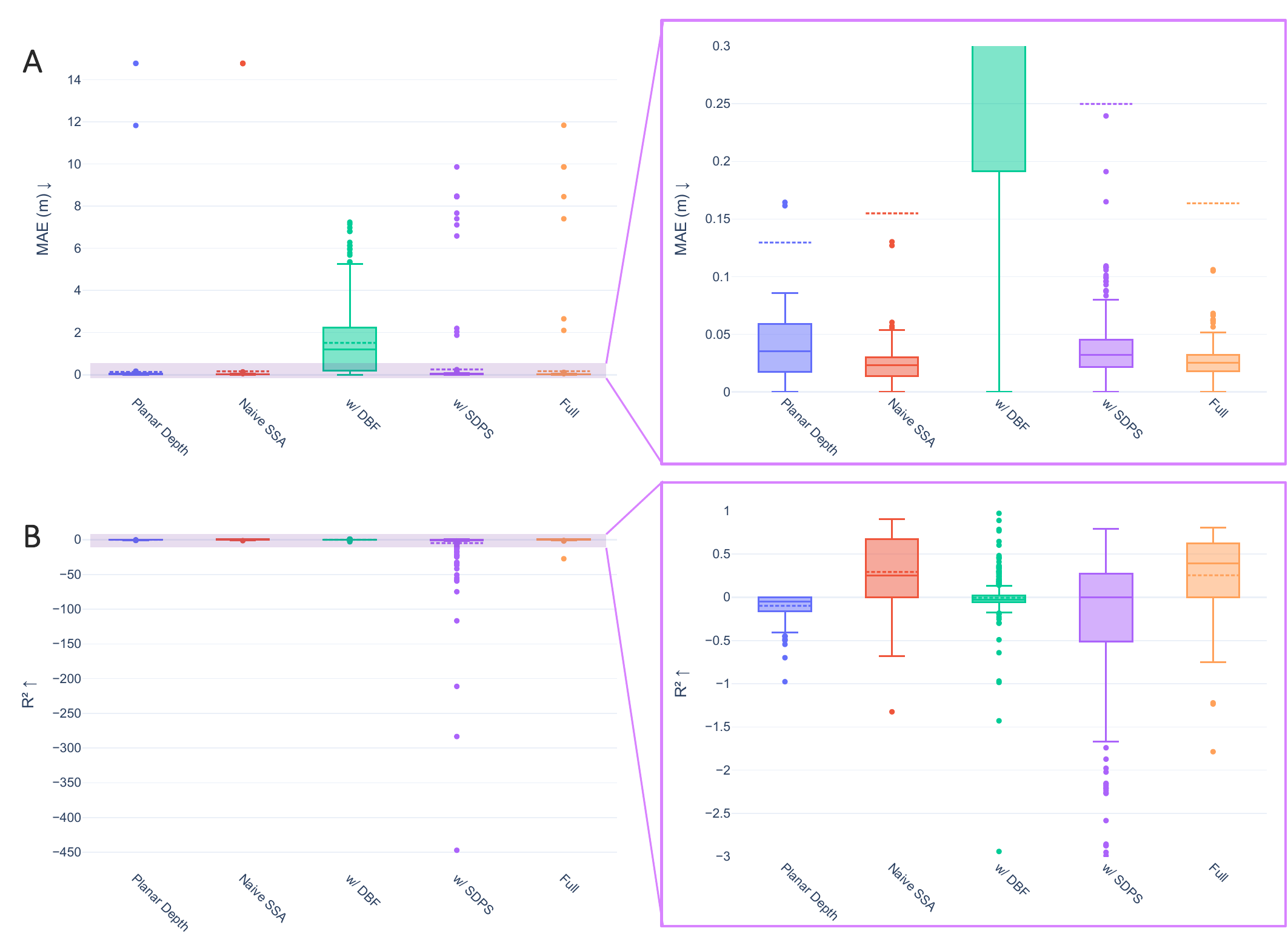}
    \caption{Distribution of per-frame evaluation metrics for the Red Hartebeest sequence. Only our Full method consistently achieves a positive $R^2$.}
    \label{fig:ablation_plots}
    \vspace{-6mm}
\end{figure}

Qualitatively, this RGB-LiDAR fusion baseline produces dense and coherent 3D point clouds that capture the overall shape and posture of the animal, preserving key anatomical features only sparsely sampled by LiDAR (Figure~\ref{fig:qualitative_results}).
Quantitatively, our analysis highlights the limitations of naive geometric models: as shown in Figure~\ref{fig:ablation_plots} and Table~\ref{tab:ablation_results}, while the Planar Depth baseline can achieve a deceptively low MAE, its $R^2$ value is consistently near or below zero, indicating that it fails to explain the variance in the ground-truth depth.
In contrast, the full RGB-LiDAR fusion baseline (``Full'') achieves a positive $R^2$, demonstrating that even a simple global alignment on top of a strong MDE prior can recover meaningful non-planar structure on WildDepth.

\begin{table}[h]
\centering
\caption{Quantitative comparison for the Red Hartebeest sequence. While MAE is comparable, only our method achieves a positive $R^2$, indicating a meaningful geometric fit.}
\label{tab:ablation_results}
\begin{tabular}{lrr}
    \toprule
    Method & MAE~($\downarrow$) & $R^2$~($\uparrow$) \\
    \midrule
    Planar Depth & $0.13 \pm 1.14$ & $-0.10 \pm 0.13$ \\
    Full (ours) & $0.16 \pm 1.12$ & $0.25 \pm 1.36$ \\
    \bottomrule
\end{tabular}
\end{table}


\section{Discussion}
\label{sec:discussion}

WildDepth introduces one of the first multimodal RGB-LiDAR datasets for long-range wildlife perception, providing a testbed for evaluating modern vision models in challenging field conditions.
Our experiments reveal several key insights.
First, large-scale monocular depth estimators such as DepthAnything transfer surprisingly well to wildlife scenes, consistently outperforming other baselines across all metrics and demonstrating the value of broad pretraining for this domain.
Second, we show that even extremely sparse LiDAR returns can be fused with dense monocular predictions to recover metrically accurate geometry.
Our RGB-LiDAR densification case study on a walking red hartebeest at nearly 100\,m confirms that a simple global alignment, anchored by LiDAR, can produce dense depth maps with positive $R^2$ scores, correctly capturing non-planar structure where naive geometric baselines fail.
Finally, our qualitative results for 4D reconstruction, which combine SAM2 with a Gaussian-splatting framework, suggest that temporally consistent 3D modeling of animal motion in the wild is becoming attainable.

However, we acknowledge several limitations that frame important directions for future work.
The current dataset is geographically restricted to Southern Africa and the UK, and does not yet include quantitative benchmarks for higher-level tasks such as pose estimation or behavior analysis, which we leave for future releases.
Our experimental validation of RGB-LiDAR fusion is based on a single, albeit challenging, sequence, and its generalization across species and environments requires further investigation.
Future work should therefore focus on three key areas: (1) expanding WildDepth's coverage to new ecosystems and species while enriching annotations for pose and behavior; (2) performing systematic cross-dataset and multimodal ablation studies to quantify domain gaps and the precise utility of LiDAR; and (3) connecting geometric outputs to downstream ecological indicators, such as body condition or population density, in collaboration with conservation practitioners.

In conclusion, WildDepth provides a new resource and a set of conservative baselines that bridge the gap between standard computer vision tasks and the realities of ecological fieldwork.
We hope this work will stimulate further research in robust multimodal perception and 4D understanding of animal behavior in the wild, ultimately contributing to global biodiversity conservation.
{\small

\bibliographystyle{ieee_fullname}
\bibliography{egbib}

@article{binley2024making,
  title={Making the most of existing data in conservation research},
  author={Binley, Allison D and Vincent, Jaimie G and Rytwinski, Trina and Soroye, Peter and Bennett, Joseph R},
  journal={Perspectives in Ecology and Conservation},
  volume={22},
  number={2},
  pages={122--128},
  year={2024},
  publisher={Elsevier}
}

@article{andualem2024channel,
  title={Channel morphological change monitoring using high-resolution LiDAR-derived DEM and multi-temporal imageries},
  author={Andualem, Tesfa Gebrie and Peters, Stefan and Hewa, Guna A and Myers, Baden R and Boland, John and Pezzaniti, David},
  journal={Science of The Total Environment},
  volume={921},
  pages={171104},
  year={2024},
  publisher={Elsevier}
}

@article{lane2007global,
  title={The global biodiversity information facility (GBIF)},
  author={Lane, Meredith A and Edwards, James L},
  journal={Systematics Association special volume},
  volume={73},
  pages={1},
  year={2007},
  publisher={London; Chapman \& Hall; 1998}
}

@article{sullivan2009ebird,
  title={eBird: A citizen-based bird observation network in the biological sciences},
  author={Sullivan, Brian L and Wood, Christopher L and Iliff, Marshall J and Bonney, Rick E and Fink, Daniel and Kelling, Steve},
  journal={Biological conservation},
  volume={142},
  number={10},
  pages={2282--2292},
  year={2009},
  publisher={Elsevier}
}

@misc{list2011iucn,
  title={IUCN red list},
  author={List, IUCN Red},
  year={2011}
}

@inproceedings{gabeff2025mammalps,
  title={MammAlps: A multi-view video behavior monitoring dataset of wild mammals in the Swiss Alps},
  author={Gabeff, Valentin and Qi, Haozhe and Flaherty, Brendan and Sumbul, Gencer and Mathis, Alexander and Tuia, Devis},
  booktitle={Proceedings of the Computer Vision and Pattern Recognition Conference},
  pages={13854--13864},
  year={2025}
}

@article{duporge2025baboonland,
  title={BaboonLand Dataset: Tracking Primates in the Wild and Automating Behaviour Recognition from Drone Videos},
  author={Duporge, Isla and Kholiavchenko, Maksim and Harel, Roi and Wolf, Scott and Rubenstein, Daniel I and Crofoot, Margaret C and Berger-Wolf, Tanya and Lee, Stephen J and Barreau, Julie and Kline, Jenna and others},
  journal={International Journal of Computer Vision},
  pages={1--12},
  year={2025},
  publisher={Springer}
}

@inproceedings{kholiavchenko2024kabr,
  title={Kabr: In-situ dataset for kenyan animal behavior recognition from drone videos},
  author={Kholiavchenko, Maksim and Kline, Jenna and Ramirez, Michelle and Stevens, Sam and Sheets, Alec and Babu, Reshma and Banerji, Namrata and Campolongo, Elizabeth and Thompson, Matthew and Van Tiel, Nina and others},
  booktitle={Proceedings of the IEEE/CVF Winter Conference on Applications of Computer Vision},
  pages={31--40},
  year={2024}
}

@inproceedings{liu2023lote,
  title={LoTE-Animal: A long time-span dataset for endangered animal behavior understanding},
  author={Liu, Dan and Hou, Jin and Huang, Shaoli and Liu, Jing and He, Yuxin and Zheng, Bochuan and Ning, Jifeng and Zhang, Jingdong},
  booktitle={Proceedings of the IEEE/CVF international conference on computer vision},
  pages={20064--20075},
  year={2023}
}

@inproceedings{yang2024depth,
  title={Depth anything: Unleashing the power of large-scale unlabeled data},
  author={Yang, Lihe and Kang, Bingyi and Huang, Zilong and Xu, Xiaogang and Feng, Jiashi and Zhao, Hengshuang},
  booktitle={Proceedings of the IEEE/CVF conference on computer vision and pattern recognition},
  pages={10371--10381},
  year={2024}
}

@article{bochkovskii2024depth,
  title={Depth pro: Sharp monocular metric depth in less than a second},
  author={Bochkovskii, Aleksei and Delaunoy, Ama{\~A}{\c{G}}l and Germain, Hugo and Santos, Marcel and Zhou, Yichao and Richter, Stephan R and Koltun, Vladlen},
  journal={arXiv preprint arXiv:2410.02073},
  year={2024}
}

@article{li2024improved,
  title={An Improved MonoDepth2 Algorithm for Vehicle Monocular Depth Estimation},
  author={Li, Chaoqun and Yue, Chenxi and Liu, Yanyan and Bie, Minglin and Li, Guoning and Lv, Zengming and Li, Jin},
  journal={Optik},
  volume={311},
  pages={171936},
  year={2024},
  publisher={Elsevier}
}

@inproceedings{ranftl2021vision,
  title={Vision transformers for dense prediction},
  author={Ranftl, Ren{\'e} and Bochkovskiy, Alexey and Koltun, Vladlen},
  booktitle={Proceedings of the IEEE/CVF international conference on computer vision},
  pages={12179--12188},
  year={2021}
}

@article{muramatsuWildPoseLongRange3D2024,
  title = {{{WildPose}}: {{A Long-Range 3D Wildlife Motion Capture System}}},
  author = {Muramatsu, Naoya and {Sangyun Shin} and {Andrew Markham} and {Amir Patel}},
  year = 2024,
  journal = {bioRxiv},
  doi = {10.1101/2024.02.05.578861},
}

@inproceedings{ng2022animal,
  title={Animal kingdom: A large and diverse dataset for animal behavior understanding},
  author={Ng, Xun Long and Ong, Kian Eng and Zheng, Qichen and Ni, Yun and Yeo, Si Yong and Liu, Jun},
  booktitle={Proceedings of the IEEE/CVF conference on computer vision and pattern recognition},
  pages={19023--19034},
  year={2022}
}

@inproceedings{chen2023mammalnet,
  title={Mammalnet: A large-scale video benchmark for mammal recognition and behavior understanding},
  author={Chen, Jun and Hu, Ming and Coker, Darren J and Berumen, Michael L and Costelloe, Blair and Beery, Sara and Rohrbach, Anna and Elhoseiny, Mohamed},
  booktitle={Proceedings of the IEEE/CVF conference on computer vision and pattern recognition},
  pages={13052--13061},
  year={2023}
}

@article{ma2023chimpact,
  title={Chimpact: A longitudinal dataset for understanding chimpanzee behaviors},
  author={Ma, Xiaoxuan and Kaufhold, Stephan and Su, Jiajun and Zhu, Wentao and Terwilliger, Jack and Meza, Andres and Zhu, Yixin and Rossano, Federico and Wang, Yizhou},
  journal={Advances in Neural Information Processing Systems},
  volume={36},
  pages={27501--27531},
  year={2023}
}

@article{rogers2023meerkat,
  title={Meerkat behaviour recognition dataset},
  author={Rogers, Mitchell and Gendron, Ga{\"e}l and Valdez, David Arturo Soriano and Azhar, Mihailo and Chen, Yang and Heidari, Shahrokh and Perelini, Caleb and O'Leary, Padriac and Knowles, Kobe and Tait, Izak and others},
  journal={arXiv preprint arXiv:2306.11326},
  year={2023}
}

@misc{wangMoGe2AccurateMonocular2025,
  title = {{{MoGe-2}}: {{Accurate Monocular Geometry}} with {{Metric Scale}} and {{Sharp Details}}},
  shorttitle = {{{MoGe-2}}},
  author = {Wang, Ruicheng and Xu, Sicheng and Dong, Yue and Deng, Yu and Xiang, Jianfeng and Lv, Zelong and Sun, Guangzhong and Tong, Xin and Yang, Jiaolong},
  year = 2025,
  month = jul,
  number = {arXiv:2507.02546},
  eprint = {2507.02546},
  primaryclass = {cs},
  publisher = {arXiv},
  doi = {10.48550/arXiv.2507.02546},
  urldate = {2025-07-06},
  archiveprefix = {arXiv}
}

@misc{raviSAM2Segment2024,
  title = {{{SAM}} 2: {{Segment Anything}} in {{Images}} and {{Videos}}},
  shorttitle = {{{SAM}} 2},
  author = {Ravi, Nikhila and Gabeur, Valentin and Hu, Yuan-Ting and Hu, Ronghang and Ryali, Chaitanya and Ma, Tengyu and Khedr, Haitham and R{\"a}dle, Roman and Rolland, Chloe and Gustafson, Laura and Mintun, Eric and Pan, Junting and Alwala, Kalyan Vasudev and Carion, Nicolas and Wu, Chao-Yuan and Girshick, Ross and Doll{\'a}r, Piotr and Feichtenhofer, Christoph},
  year = 2024,
  month = aug,
  number = {arXiv:2408.00714},
  eprint = {2408.00714},
  primaryclass = {cs},
  publisher = {arXiv},
  doi = {10.48550/arXiv.2408.00714},
  urldate = {2024-08-19},
  archiveprefix = {arXiv},
}

@inproceedings{feng2025gaussian,
  title={Gaussian splashing: Unified particles for versatile motion synthesis and rendering},
  author={Feng, Yutao and Feng, Xiang and Shang, Yintong and Jiang, Ying and Yu, Chang and Zong, Zeshun and Shao, Tianjia and Wu, Hongzhi and Zhou, Kun and Jiang, Chenfanfu and others},
  booktitle={Proceedings of the Computer Vision and Pattern Recognition Conference},
  pages={518--529},
  year={2025}
}

@article{macenski2022robot,
  title={Robot operating system 2: Design, architecture, and uses in the wild},
  author={Macenski, Steven and Foote, Tully and Gerkey, Brian and Lalancette, Chris and Woodall, William},
  journal={Science robotics},
  volume={7},
  number={66},
  pages={eabm6074},
  year={2022},
  publisher={American Association for the Advancement of Science}
}

@article{ionescu2013human3,
  title={Human3. 6m: Large scale datasets and predictive methods for 3d human sensing in natural environments},
  author={Ionescu, Catalin and Papava, Dragos and Olaru, Vlad and Sminchisescu, Cristian},
  journal={IEEE transactions on pattern analysis and machine intelligence},
  volume={36},
  number={7},
  pages={1325--1339},
  year={2013},
  publisher={IEEE}
}

@inproceedings{mahmood2019amass,
  title={AMASS: Archive of motion capture as surface shapes},
  author={Mahmood, Naureen and Ghorbani, Nima and Troje, Nikolaus F and Pons-Moll, Gerard and Black, Michael J},
  booktitle={Proceedings of the IEEE/CVF international conference on computer vision},
  pages={5442--5451},
  year={2019}
}

@article{ye2024superanimal,
  title={SuperAnimal pretrained pose estimation models for behavioral analysis},
  author={Ye, Shaokai and Filippova, Anastasiia and Lauer, Jessy and Schneider, Steffen and Vidal, Maxime and Qiu, Tian and Mathis, Alexander and Mathis, Mackenzie Weygandt},
  journal={Nature communications},
  volume={15},
  number={1},
  pages={5165},
  year={2024},
  publisher={Nature Publishing Group UK London}
}

@article{xiao2020audiovisual,
  title={Audiovisual slowfast networks for video recognition},
  author={Xiao, Fanyi and Lee, Yong Jae and Grauman, Kristen and Malik, Jitendra and Feichtenhofer, Christoph},
  journal={arXiv preprint arXiv:2001.08740},
  year={2020}
}

@article{liao2022kitti,
  title={Kitti-360: A novel dataset and benchmarks for urban scene understanding in 2d and 3d},
  author={Liao, Yiyi and Xie, Jun and Geiger, Andreas},
  journal={IEEE Transactions on Pattern Analysis and Machine Intelligence},
  volume={45},
  number={3},
  pages={3292--3310},
  year={2022},
  publisher={IEEE}
}

@inproceedings{caesar2020nuscenes,
  title={nuscenes: A multimodal dataset for autonomous driving},
  author={Caesar, Holger and Bankiti, Varun and Lang, Alex H and Vora, Sourabh and Liong, Venice Erin and Xu, Qiang and Krishnan, Anush and Pan, Yu and Baldan, Giancarlo and Beijbom, Oscar},
  booktitle={Proceedings of the IEEE/CVF conference on computer vision and pattern recognition},
  pages={11621--11631},
  year={2020}
}

@inproceedings{silberman2012indoor,
  title={Indoor segmentation and support inference from rgbd images},
  author={Silberman, Nathan and Hoiem, Derek and Kohli, Pushmeet and Fergus, Rob},
  booktitle={European conference on computer vision},
  pages={746--760},
  year={2012},
  organization={Springer}
}

@article{ravoor2020deep,
  title={Deep learning methods for multi-species animal re-identification and tracking--a survey},
  author={Ravoor, Prashanth C and Sudarshan, TSB},
  journal={Computer Science Review},
  volume={38},
  pages={100289},
  year={2020},
  publisher={Elsevier}
}

@article{beeryMegadetector2019,
  author       = {Sara Beery and
                  Dan Morris and
                  Siyu Yang},
  title        = {Efficient Pipeline for Camera Trap Image Review},
  journal      = {CoRR},
  volume       = {abs/1907.06772},
  year         = {2019},
  url          = {http://arxiv.org/abs/1907.06772},
  eprinttype    = {arXiv},
  eprint       = {1907.06772},
  bibsource    = {dblp computer science bibliography, https://dblp.org}
}

@inproceedings{zhang2022bytetrack,
  title={Bytetrack: Multi-object tracking by associating every detection box},
  author={Zhang, Yifu and Sun, Peize and Jiang, Yi and Yu, Dongdong and Weng, Fucheng and Yuan, Zehuan and Luo, Ping and Liu, Wenyu and Wang, Xinggang},
  booktitle={European conference on computer vision},
  pages={1--21},
  year={2022},
  organization={Springer}
}

@inproceedings{wang2023videomae,
  title={Videomae v2: Scaling video masked autoencoders with dual masking},
  author={Wang, Limin and Huang, Bingkun and Zhao, Zhiyu and Tong, Zhan and He, Yinan and Wang, Yi and Wang, Yali and Qiao, Yu},
  booktitle={Proceedings of the IEEE/CVF conference on computer vision and pattern recognition},
  pages={14549--14560},
  year={2023}
}
}

\end{document}